\documentclass{article}

\usepackage[utf8]{inputenc} 
\usepackage{graphicx}
\usepackage{multirow}
\usepackage[lofdepth,lotdepth]{subfig}
\usepackage{multicol}
\usepackage{amsmath}
\usepackage{threeparttablex}
\usepackage{enumitem}
\usepackage{pifont}
\usepackage{rotating}
\usepackage{authblk}

\newcommand{\cmark}{\ding{51}}

\begin{document}



\title{Detecting Malicious Accounts in Permissionless Blockchains using Temporal Graph Properties}

\author[1]{Rachit Agarwal}
\author[1]{Shikhar Barve}
\author[1]{Sandeep K. Shukla}
\affil[1]{CSE Department, IIT Kanpur, India}
\affil[1]{\{rachitag, shikhar, sandeeps\}@cse.iitk.ac.in}

\date{}                    



\providecommand{\keywords}[1]{\textbf{\textit{Index terms---}} #1}

\maketitle

\begin{abstract} 
The temporal nature of modeling accounts as nodes and transactions as directed edges in a directed graph -- for a blockchain,  enables us to understand the behavior (malicious or benign) of the accounts. Predictive classification  of accounts as malicious or benign  could help users of the permissionless blockchain platforms to operate in a secure manner. Motivated by this, we introduce temporal features such as burst and attractiveness on top of several already used graph properties such as the node degree and clustering coefficient. Using identified features, we train various Machine Learning (ML) algorithms and identify the algorithm that performs the best in detecting which accounts are malicious. We then study the behavior of the accounts over different temporal granularities of the dataset before assigning them malicious tags. For Ethereum blockchain, we identify that for the entire dataset - the ExtraTreesClassifier performs the best among supervised ML algorithms. On the other hand, using cosine similarity on top of the results provided by unsupervised ML algorithms such as K-Means on the entire dataset, we were able to detect 554 more suspicious accounts. Further, using behavior change analysis for accounts, we identify 814 unique suspicious accounts across different temporal granularities.

\end{abstract}




\section{Introduction}\label{sec:intro}

A Blockchains is an ever-growing large directed temporal network with more and more industries starting to adopt it for their businesses. In permissionless blockchains, interactions (also called as transactions) happen between different types of accounts. In Ethereum mainnet public blockchain, these accounts can be either Externally Owned Accounts (\textit{EOA}) or Smart Contracts (\textit{SC}). Here, transactions from an EOA (called as an \textit{external transaction}) are  recorded on the blockchain ledger whereas transactions from an SC (called as an \textit{internal transaction}) are  not recorded on the ledger.

With actual money involved in most of the permissionless blockchains, an account must be able to perform secure transactions. Recently, many security threats to various blockchain platforms have been identified~\cite{Bryk2018}. For some identified vulnerabilities, counter-measures have been implemented. We do not delve into surveying all the security threats. In \cite{chen2019survey}, the authors survey security flaws that exist in Ethereum blockchains. In many of the security vulnerabilities identified in Ethereum blockchain, hackers target other accounts by either hacking SCs or implementing malicious SCs for cybercrimes such as ransomware, scams, phishing, and hacking of exchanges or wallets~\cite{ChainAnalysis2019}. 

With an ever-increasing growth and adoption of blockchain technology by the industry and the crypto-currency market, permissionless blockchains are at the epicenter of increased security vulnerabilities and attacks. Our motivation for this work is based on the fact that there is limited work on learning the behaviors of the accounts in permissionless blockchains which are malicious and potentially victimize other accounts in the future. In short, we aim to identify malicious accounts so that the potential victims and blockchains can deploy counter-measures. In this paper, henceforth, we use term blockchain to represent permissionless blockchain. The techniques proposed in related studies classify accounts as malicious using either machine learning (ML) algorithms or motif-based (basic building subgraphs of a network) methods.  Nonetheless, the features used by the available techniques are: (\textit{a}) limited and not learned from the previous attacks on blockchains, and (\textit{b}) extracted from the aggregated snapshot of time-dependent transaction graphs that do not consider temporal evolution of the graphs.  

The temporal aspects attached to the features are essential in understanding the actual behavior of an account before we can classify it as malicious. For example, inDegree and outDegree features are time-variant and should be considered a time series. Nonetheless, it has been proven that the aggregated node degree distribution for accounts follows a power-law in blockchains such as Ethereum~\cite{chen2018}. Here, questions that we ask are: \textit{does such behavior exist in all accounts? Is there a burst of degree for certain accounts at certain instances and can the existence of such bursts be used to identify malicious activity?} To answer these question, we first identify the existence of bursts. Then to study the effect of bursts, we introduce features such as \textit{\textbf{temporal burst}}, \textit{\textbf{degree burst}}, \textit{\textbf{balance burst}}, and \textit{\textbf{gasPrice burst}}. 

The fat-tailed nature of power-law degree distribution also gives rise to neighbor-hood-based fitness preferential attachment in blockchains~\cite{Aspembitova2019}.  In~\cite{Aspembitova2019} authors defined fitness as ``\textit{the ability of the node to attract new connections}" and showed that the accounts that have high fitness sometimes are short-lived and indulge mostly in malicious activities while when they are long-lived they represent large organizations. Here, the authors define the fitness factor considering one previous time instance interactions. As it does not consider a temporal window, one drawback of the method lies in its ability to correctly classify malicious transactions that appear at an interval of 2 time units or more. Inspired by this, we define a neighborhood-based feature called \textit{\textbf{attractiveness}} that takes into account a temporal window of size $\theta_a$ where $(0< \theta_a < T_{DS})$ and $T_{DS}$ is the duration for which we collect the dataset ($DS$). Our attractiveness measure takes into account the stability of directed transactions that happened between two accounts in the past. Intuitively, a malicious account will have high attractiveness as it will tend to transact with new accounts while benign accounts will have high neighborhood stability or low attractiveness. 

As the behavior of an account can change from malicious to benign or from benign to malicious over time, there is a need for continuous monitoring and analysis of the real-time transactions given the history of transactions performed by an account. We thus study the evolution of malicious behavior over different timescales by creating sub datasets and then answer \textit{would a certain account show malicious behavior in future?} Towards this, we first apply different ML algorithms and identify the most suitable unsupervised ML algorithm in the entire dataset that is able to cluster accounts most accurately. Then we apply the identified algorithm to different sub datasets within a temporal scale to capture the behavior changes.

In summary, following are our main contributions:
\begin{itemize}[leftmargin=*]
    \item \textit{\textbf{Feature Engineering}}: We identify \textbf{feature vector} for identifying malicious accounts based on \textbf{previous attacks on blockchains} and perform time series analysis. As new features, we propose \textit{\textbf{temporal burst}}, \textit{\textbf{degree burst}}, \textit{\textbf{balance burst}}, \textit{\textbf{gasPrice burst}}, and \textit{\textbf{attractiveness}}.  
    
    \item \textit{\textbf{Comparative analysis}}: We perform a \textbf{comparative study} with techniques proposed in related studies and identify best possible \textbf{supervised and unsupervised ML algorithm} with related hyperparameters when we use Ethereum transaction data.
    
    \item \textit{\textbf{Results}}: Our results demonstrate that \textbf{ExtraTreesClassifier} performs best with respect to balanced accuracy under supervised settings for the entire dataset while when using clustering techniques, we are able to identify 554 more suspect accounts. Analysis of behavioral changes reveal 814 suspects across different temporal granularities.
    
\end{itemize}

The rest of the paper is organized as follows. In section~\ref{sec:rw}, we present background and the state of the art techniques for identifying  malicious accounts and compare them. In sections~\ref{sec:method} and section~\ref{sec:fe}, we present detailed description of our methodology and the feature vector, respectively. This is followed by in-depth evaluation along with the results in section~\ref{sec:eval}. We finally conclude in section~\ref{sec:conclusion} providing details on prospective future work. Further, in Table~\ref{tab:acronym} we provide list of acronyms used in the paper.

\begin{table}
\centering
\begin{tabular}{|c|l| | c|l|}
\hline
\textbf{Acronym} & \textbf{Meaning} & \textbf{Acronym} & \textbf{Meaning}\\
\hline
EOA & Externally Owned Account & CC & Clustering Coefficient\\
SC & Smart Contract & Bal & Balance\\
ML & Machine Learning & TF & Transaction Fee\\
AS & Active State & BB & Burst\\
iD & inDegree & IET & Inter-Event Time\\
oD & outDegree & A & Attractiveness\\
PoW & Proof of Work &LOF & Local Outlier Factor\\
EVM & Ethereum Virtual Machine &&\\
\hline
\end{tabular}
\caption{List of Acronyms.}
\label{tab:acronym}
\end{table}

\section{Background and Related Work}\label{sec:rw}

There are two types of blockchain technologies, permissionless and permissioned. The major difference between two technologies is that in permissioned blockchain prior access approval is needed for performing any action on the blockchain while in permissionless blockchain anyone can perform actions on the blockchain without any approval. Further, there is no way to censor anyone from permissionless blockchains. Such aspects lead to more frauds and malicious activities to prevail in permissionless blockchains. Ethereum and Bitcoin use permissionless technology.

Ethereum was developed by Vitalik Buterin in 2013~\cite{Vitalik2018} and allows users to run programs in its trusted virtual environment known as Ethereum Virtual Machine (EVM). These programs are called Smart Contracts (SC) and are stored on the ledger along with transactions performed on a given fixed address. Ethereum uses ``\textit{Ether}'' as its native crypto-currency for transfer and transaction fees. Smart Contracts can also send, store and receive ethers. Once deployed it is a hard coded program that could only be fed with input to get output. Smart Contracts are also used by some applications for their processing. Such applications are called distributed applications or dapps. Although Ethereum is known for its security and trust a small bug in SC code can cause huge loss~\cite{Atzei2017} of crypto-currency. 
Unlike Bitcoin, Ethereum uses list of accounts. For a valid transaction, amount is transferred from sender to receiver. If receiver is a SC, its code is executed and the state of the SC is updated. Internally, a SC could send a message or perform internal transactions with other accounts. Ethereum currently uses a refined form of PoW (Proof of Work) consensus algorithm. PoW is computationally expensive and energy inefficient. 

There are vast number of studies in fraud detection~\cite{abdallah2016}. Nonetheless, targeting Ethereum, Chen et al.~\cite{chen2019survey} base their survey on attacks and defences in Ethereum. We do not survey all the attacks and defense mechanisms in this work. However, we provide an in-depth understanding of different methods used to detect accounts involved in malicious activity. Several works have tried to identify or categorize malicious accounts and activities in different types of blockchains. As blockchains have graph structure, most of these techniques study graph properties (such as node degree) to identify features before applying supervised or unsupervised learning. 

In~\cite{pham2016anomaly}, authors used a bitcoin transaction network to detect malicious activity. They were able to detect three malicious attacks using unsupervised ML algorithms with a limited amount of available transaction data. In their followup work, they used a more comprehensive bitcoin transaction dataset (starting from genesis block until April 7$^{th}$, 2013)~\cite{pham2017anomaly}. They employed data in two types of graphs namely \textit{User Graph} and \textit{Transaction Graph}. In user-graph nodes represent accounts and edges represent transactions, whereas in transaction-graph nodes represent transactions and edges represent flow of bitcoins. They first studied the flow of bitcoins to prove the existence of anomalies and then performed clustering to identify different attacks. They were able to detect the existence of one attack using the \textit{Local Outlier Factor} (LOF). Inspired by~\cite{pham2017anomaly}, in~\cite{monamo2016}, Monamo et al. also used bitcoin transaction data and proposed an update to counter scaling issues that are inherent in LOF. They validated their approach using \textit{trimmed K-Means}, argued its usefulness in detecting anomalies and  detected 5 out of 30 fraudsters. 

In another bitcoin-related malicious activity detection~\cite{Bartoletti2018}, authors studied the detection of addresses involved in the Ponzi scheme. They used supervised learning and validated their results after addressing the class imbalance that is inherent in any malicious activity related to datasets. They identified that the Gini coefficient of outgoing values and the ratio between incoming and total transactions are the most important features for detecting Ponzi scheme related accounts. In another Ponzi scheme related study, in~\cite{Chen2018WWW}, authors use Ethereum data to extract features from operation codes (opcodes) of the smart contract's bytecode. Their motivation behind the study was based on the fact that the opcodes reflect logic implemented in a SC and therefore provide useful features for identifying Ponzi and non-Ponzi SC. They also figured out that opcode features are more efficient than account based features while detecting Ponzi scheme accounts.  In~\cite{ostapowicz2019detecting}, authors use partial Ethereum transaction data to classify malicious accounts. They also performed a sensitivity analysis to study the effect of different classifiers on the feature set. In~\cite{ajay2019}, to counter class imbalance, authors assumed that accounts connected to malicious accounts via incoming transactions are also malicious. They then studied various supervised ML algorithms to identify malicious accounts over this over-sampled Ethereum dataset. In a followup study of~\cite{ajay2019}, in~\cite{kumar2020}, authors used only those benign accounts who have never transacted with malicious accounts. Due to this, their feature vector has only transaction based properties but not the graph based properties. 

\textit{N-motifs} are frequently occurring subgraphs that serve as a basic building block of a network. Authors in \cite{zola2019ieee} defined N-motif as a path of length $2N$ between two entities where transactions are also considered as vertices. Using N-motifs that are present in the transaction graph, in~\cite{zola2019ieee}, authors studied transactions happening between entities (people or organizations with multiple accounts). They were able to correctly identify malicious accounts involved in gambling. In another study~\cite{Goldsmith2020}, authors analysed transfer of funds within a subnet and used temporal feature such as how quickly funds are cashed.

We present all the above-mentioned techniques in detail in Table~\ref{table:rw} and present the features that the techniques used along with studied ML algorithm, their hyperparameters, accounts considered in the dataset and performance score. Note that all these techniques use features that are based on some graph properties, transacting amount, and active state to train the ML model. However, several other studies, such as~\cite{chen2018,cheng2019}, use inferences drawn from the analysis of the transaction graph to mark malicious accounts. In~\cite{chen2018}, authors try to identify accounts indulging in DDos attack and argue that accounts that create multiple rarely used contracts are malicious. A similar approach is followed in~\cite{Jung2019} where they used only verified SC codes and introduced features like SC size, lifetime and average time between transactions (i.e. Inter-event time). In~\cite{cheng2019}, authors deploy honeypot and analyze RPC requests to identify malicious accounts. They then analyze transactions to mark accounts as suspicious that accept crypto-currency from malicious accounts. They perform behavior analysis to identify \textit{fisher} accounts and attacks such as crypto-currency stealing.

\begin{sidewaystable}
\centering
\small
\caption{Features used in related studies}
\label{table:rw}
\begin{threeparttable}
\begin{tabular}{|c|c|ccccccccc|c|c|c|c|}
\hline
&  & \multicolumn{9}{c|}{\textbf{Used features based on}}  &  & &  & \\
\hline
\textbf{\#} & \textbf{B/C} & \textbf{AS} & \textbf{iD}  & \textbf{oD} & \textbf{Bal}   & \textbf{TF} & \textbf{BB} & \textbf{A} & \textbf{CC} & \textbf{IET} & \textbf{ML Algo Used} & \textbf{Dataset} & \textbf{Hyperparameters} &  \textbf{Performance} \\
\hline

\multirow{3}{*}{\cite{pham2016anomaly}} & \multirow{3}{*}{B} & \multirow{3}{*}{\cmark} & \multirow{3}{*}{\cmark} & \multirow{3}{*}{\cmark} & \multirow{3}{*}{\cmark} & \multirow{3}{*}{-} & \multirow{3}{*}{-} & \multirow{3}{*}{-} & \multirow{3}{*}{\cmark} & \multirow{3}{*}{\cmark} & K-Means & \multirow{3}{*}{100K$^a$} & $k \in[1,14]$ & $k_{opt}=7,8$ \\
\cline{12-12} \cline{14-15} 
& & & & & & & & & & & Mahalanobis Distance & & $\times$ & 0.0256$^{MDE}$ \\ 
\cline{12-12} \cline{14-15} 
& & & & & & & & &  & & $\nu$-SVM & & $\nu=0.005$ & 0.1441$^{MDE}$ \\
\hline
\cite{pham2017anomaly} & B & \cmark & \cmark  & \cmark & \cmark  & - & -  & - & -  & \cmark  & Local Outlier Factor & 6.3M$^a$ &  $k = 8$ & 0.55$^{MDE}$ \\ 
\hline
\multirow{2}{*}{\cite{monamo2016}} & \multirow{2}{*}{B} & \multirow{2}{*}{-} & \multirow{2}{*}{\cmark} & \multirow{2}{*}{\cmark} & \multirow{2}{*}{\cmark} & \multirow{2}{*}{-} & \multirow{2}{*}{-} & \multirow{2}{*}{-} & \multirow{2}{*}{\cmark}  & \multirow{2}{*}{-} & K-Means & \multirow{2}{*}{1M$^a$} & $k \in [1,14]$ & $k_{opt}=8$ \\ 
\cline{12-12} \cline{14-15} 
& & & & & & & & & & & Trimmed K-Means & & $k \in [1,15]$, $\alpha=0.01$ & $k_{opt}=8$ \\ 
\hline
\multirow{3}{*}{\cite{Bartoletti2018}} & \multirow{3}{*}{B} & \multirow{3}{*}{\cmark} & \multirow{3}{*}{\cmark} & \multirow{3}{*}{\cmark} & \multirow{3}{*}{\cmark} & \multirow{3}{*}{-} & \multirow{3}{*}{-} & \multirow{3}{*}{-} & \multirow{3}{*}{-} & \multirow{3}{*}{\cmark} & RIPPER$\dagger$ & \multirow{3}{*}{$\ddagger$6432$^a$} & cost $\in[1,40]$ & 0.996$^{ac}$ \\ 
\cline{12-12} \cline{14-15} 
 & & & & & & & & & & & Bayes Network & & $\times$ & 0.983$^{ac}$ \\ 
\cline{12-12} \cline{14-15} 
& & & & & & & & & & & Random Forest & & $\times$ & 0.996$^{ac}$ \\
\hline
\cite{Chen2018WWW} & E & - & \cmark & \cmark & \cmark & \cmark & - & - & - & - & XGBoost & $\ddagger$1382$^{sc}$ & $\times$ & 0.94$^p$, 0.81$^r$ \\ 
\hline
\multirow{3}{*}{\cite{ostapowicz2019detecting}} & \multirow{3}{*}{E} & \multirow{3}{*}{\cmark} & \multirow{3}{*}{\cmark} & \multirow{3}{*}{\cmark} & \multirow{3}{*}{-}  & \multirow{3}{*}{\cmark} & \multirow{3}{*}{-} & \multirow{3}{*}{-} & \multirow{3}{*}{-} & \multirow{3}{*}{-} & Random Forest & \multirow{3}{*}{350K$^a$} & RFPARAM & 0.85$^{r}$, 0.05$^{p}$ \\
\cline{12-12} \cline{14-15} 
& & & & & & & & & & & SVM & & $cost=1$, $\gamma=0.077$ & 0.87$^{r}$, 0.02$^{p}$ \\ 
\cline{12-12} \cline{14-15} 
& & & & & & & & & & & XGBoost  & & XGBPARAM & 0.8$^{r}$, 0.07$^{p}$  \\ \hline
\multirow{6}{*}{\cite{ajay2019}} & \multirow{6}{*}{E} & \multirow{6}{*}{\cmark} & \multirow{6}{*}{\cmark} & \multirow{6}{*}{\cmark} & \multirow{6}{*}{\cmark} & \multirow{6}{*}{-} & \multirow{6}{*}{-} & \multirow{6}{*}{-} & \multirow{6}{*}{-} & \multirow{6}{*}{-} & Decision Tree & \multirow{6}{*}{300$^a$} & $\times$ & 0.93$^{ac}$\\
\cline{12-12} \cline{14-15} 
& & & & & & & & & & & SVM &  & $\times$ & 0.83$^{ac}$ \\ 
\cline{12-12} \cline{14-15} 
& & & & & & & & & & & KNN & & $k=5$  & 0.91$^{ac}$ \\ 
\cline{12-12} \cline{14-15} 
& & & & & & & & & & & MLP & & $\times$ & 0.86$^{ac}$ \\ 
\cline{12-12} \cline{14-15} 
& & & & & & & & & & & NaiveBayes & &$\times$& 0.89$^{ac}$ \\
\cline{12-12} \cline{14-15} 
& & & & & & & & & & & Random Forest & & $\times$ & 0.99$^{ac}$ \\
\hline

\multirow{4}{*}{\cite{kumar2020}} & \multirow{4}{*}{E} & \multirow{4}{*}{\cmark} & \multirow{4}{*}{-} & \multirow{4}{*}{-} & \multirow{4}{*}{\cmark} & \multirow{4}{*}{\cmark} & \multirow{4}{*}{-} & \multirow{4}{*}{-} & \multirow{4}{*}{-} & \multirow{4}{*}{-} & Decision Tree & \multirow{4}{*}{9375$^a$} & $\times$ & 0.92$^{ac}$\\
\cline{12-12} \cline{14-15} 
& & & & & & & & & & & KNN & & $\times$  & 0.92$^{ac}$ \\ 
\cline{12-12} \cline{14-15} 
& & & & & & & & & & & XGBoost & &$\times$& 0.96$^{ac}$ \\
\cline{12-12} \cline{14-15} 
& & & & & & & & & & & Random Forest & & $\times$ & 0.95$^{ac}$ \\
\hline

\multirow{4}{*}{\cite{zola2019ieee}} & \multirow{4}{*}{B} & \multirow{4}{*}{-} & \multirow{4}{*}{\cmark} & \multirow{4}{*}{\cmark} & \multirow{4}{*}{\cmark} & \multirow{4}{*}{\cmark} & \multirow{4}{*}{-} & \multirow{4}{*}{-} & \multirow{4}{*}{-} & \multirow{4}{*}{-} & Adaboost & \multirow{4}{*}{1000M$^a$} & $estimators=50$, $rate=1$ & $>0.2^{r}$\\
\cline{12-12} \cline{14-15} 
& & & & & & & & & & & Random Forest &  & $estimators=10$ & $>0.85^{r}$ \\ 
\cline{12-12} \cline{14-15} 
& & & & & & & & & & & \multirow{2}{*}{Gradient boosting} & & $estimators=100$, $rate=0.1$  & \multirow{2}{*}{$>0.93^{r}$} \\ 
& & & & & & & & & & & & & $depth=3$  & \\ 
\cline{12-12} \cline{14-15} 
\hline
\end{tabular}
\begin{tablenotes}
\item $^{B/C}$ Blockchain, $^B$ Bitcoin, $^E$ Ethereum, $^a$ accounts, $^{sc}$ smart contracts, $^{MDE}$ Dual Evaluation Metric, $^{ac}$ accuracy, $^p$ Precision, $^r$ Recall, $^\dagger$ it is a propositional rule learner that relies on a sequential covering logic, $^\ddagger$ Ponzi scheme data, $^{RFPARM}$ features = 3, leaf samples = 10, threshold probability = 0.99, $^{XGBPARAM}$ depth = 3, child weight = 8, subsample = 1, probability = 0.99, $^\times$ not provided.
\end{tablenotes}
\end{threeparttable}
\end{sidewaystable}

All the above techniques either use a limited set of ML algorithms on a highly scaled-down data inducing over-fitting or apply inferences on the graph structure to identify malicious activities and accounts. In most cases, studies use features that do not capture temporal behavior and are approximated by the mean behavior, thereby, further inducing a bias in their study thus having high accuracy. Techniques that use large datasets and have high class imbalance, on the other hand, either have high recall and low precision or low recall and high precision~\cite{ostapowicz2019detecting}. Nonetheless, using our features, we identified ML algorithm that provides better precision as well as better recall.

\section{Methodology}\label{sec:method}

We use \textit{Ethereum mainnet blockchain transaction data} and first validate our assumptions and approach. We segment the transaction data into sub-datasets ($SD$) to capture the behavioral changes. We create the $SD$s using different temporal granularities ($T_g$ such that $T_g\in T_G$) where $T_G=\{Day$, $Week$, $Month$, $Quarter$, $HalfYearly$, $Year$, $All\}$. A granularity becomes coarser as we move from Day to Year. Here a $SD$ in a \textit{Day} consists of transactions of 6000 blocks. The choice of 6000 blocks is based on the fact that in Ethereum approx 6000 blocks are created every day. At a coarser $T_g$, a $SD$ in a \textit{Week} consists of 7 Days data. Similarly, a $SD$ in a \textit{Month} consists of 30 Days data, a $SD$ in a \textit{Quarter} consists of 3 Month data, a $SD$ in a \textit{HalfYearly} consists of 6 Months data, and a $SD$ in a \textit{Year} consists of 12 Months of data.  

On all the features that are time series based (features described in section~\ref{sec:fe}), we perform time series analysis of all the $SD$s at different $T_g$ to quantify them using \textit{tsfresh} that ``\textit{extracts characteristics from time series}''~\cite{christ2016,christ2018}. The analysis reveals that features such as quantile and median best describe the time series for most of the features we have. We observe this behavior not only in the entire dataset but also in different $SD$s at different $T_g$s.

We first apply the AutoML pipeline using TPOT~\cite{OlsonGECCO2016} to identify the best ML classifier on the entire dataset and validate state of the art techniques. We configure TPOT with existing tested ML algorithms and their hyperparameters. Note that TPOT internally performs imputation and feature scaling also. Nonetheless, as our aim is to detect malicious accounts, we also apply clustering to identify accounts that show similar behavior to that of malicious accounts. For the entire dataset, we find that K-Means provides best silhouette score for $k=9$ when we consider both EOAs and SCs. For clusters identified as malicious, we use cosine similarity to quantify the similarity among the accounts within the cluster. We acknowledge that there are other methods as well to identify similarity, but for this work we use cosine similarity. With this method we are able to identify 293 more suspect accounts that have similar behavior as malicious accounts. When considering only EOAs, we identify best silhouette score at $k=10$ and 554 more suspects.

Assuming that K-Means with hyperparameter $k=9$ identified for entire dataset performs best for all temporal sub-datasets at different temporal granularities, we determine a probability for an account to be malicious at different temporal granularities. Across all temporal granularities we identify 814 unique accounts as suspects.

\section{Feature Engineering}\label{sec:fe}

We do not describe the blockchain graph models as they are well understood. Instead, we directly present features that we extract from the blockchain temporal graph structure. The set of features ($F$) defined in the related work is limited and, in most cases, does not convey correct temporal behavior. We extend the feature set and introduce new features to detect malicious accounts. We follow a two-fold methodology to identify the relevant features. First, we study different attacks that have happened in the past to understand what features malicious accounts have used for malicious activity.
Second, as most of the account features (for example, inDegree) are time series based, we perform time series analysis to identify features that best represent the salient properties  of  the relevant  time series. Below we provide a list of all the features we use:

\begin{itemize}[leftmargin=*]
    \item \textit{\textbf{Non Time Series based (set $F_n | F_n\subset F$)}}
        \begin{itemize}[leftmargin=*]
            \item \textit{\textbf{Active state (AS)}}: malicious activities are usually short-lived~\cite{Aspembitova2019} and remain, for example, until remediation is   introduced. It is thus essential that we consider features such as when the account first transacted (\textit{\textbf{transactedFirst}}), last transacted (\textit{\textbf{transactedLast}}), how long it has been active (\textit{\textbf{durationActive}}), and since when the account is continuously transacting (\textit{\textbf{activeSinceLast}}).
        \end{itemize}
        
    \item \textit{\textbf{Time Series based (set $F_t | F_t\subset F$)}}: We analyze each of the following time series based features using \textit{tsfresh}~\cite{christ2016,christ2018} and select 3 top features identified for each of the following attributes. Nonetheless, as \textit{\textbf{inter-event time (IET)}} itself is a time series, we use it as a feature as well.
    
    \begin{itemize}[leftmargin=*]
        \item \textit{\textbf{inDegree (iD)}}: it represents the number of transactions in which the account under consideration is a receiver at a particular instant. Most of the malicious activities involve transfer of money to a malicious account. Thus, it is one of the most important features used to understand the behavior of a malicious account. In~\cite{ajay2019}, the author found that \textbf{\textit{uniqueInDegree}} (defined as unique accounts from which the account under consideration has ever received money) to be one of the most critical feature for identification of malicious accounts. On top, we also use aggregated \textit{\textbf{inDegreeAgg}} as a feature.
        
        \item \textit{\textbf{outDegree (oD)}}: represents the  number of transactions in which the account under consideration has sent money at a particular instant. In some attacks such as Bitpoint Hack~\cite{cointelegraph}, after the attacker has received amount of sum from the victims in an alias account, they transferred the received sum to another account they hold or to an exchange. Such attacks increase the importance of outDegree as a potential feature. Similar to the case above, we also use aggregated \textit{\textbf{outDegreeAgg}} as a feature. 
     
        \item \textit{\textbf{Balance (Bal)}}: our motivation to use it as a feature is based on the fact that most malicious activities in a permissionless blockchains are finance based. For example, in one of the famous Parity Multisig wallets~\cite{openzeppelin} attack the malicious account drained more than 150k Ethers (currency used in Ethereum blockchain). Thus the currency held by an account as well as its flow is an important feature. We identify balance time series for both in/out case. Besides balance, we identify for each instance max balance for both in and out cases (\textit{\textbf{maxInBalance}} and \textit{\textbf{maxOutBalance}}), \textit{\textbf{zeroBalanceTransactions}} (transactions where no money was transferred either to or from an account), \textit{\textbf{totalBalance}} (final balance held with the account), and \textit{\textbf{averagePerInBalance}} (average of received balance) as features. 
        
        \item \textit{\textbf{Transaction Fees (TF)}}: in crypto-currency based blockchains, a transaction is marked by \textit{transaction fees} that a sender is willing to spend on a particular transaction. In Ethereum blockchain, operations like transferring Ethers require a fixed sequence of instructions which consume 21,000 Gas ($TF=Gas\times GasPrice$). Several attackers put higher gas price to bribe the miner so that a particular transaction of interest to them is included in the next block~\cite{cheng2019}. Nonetheless, in DDos attack~\cite{spamAttack}, an attacker created multiple accounts at very low gas price to increase synchronization and processing time. Thus it is also an essential features.
                
        \item \textit{\textbf{Attractiveness (A)}}: mostly, malicious accounts tend to interact with accounts that they have not interacted with before. The probability of interacting with the same account that they have interacted before is very low. Consider $N_i^t$ to be the neighborhood (accounts with whom the account $i$ has received crypto-currency) of account $i$ at time $t$, $T=\{t, t-1, \cdots, t-\theta_a\}$, and $\theta_a$ the time window size. Based on this, we define \textit{\textbf{attractiveness}} ($A_i^t$) for account $i$ at time $t$ as shown in equation~\ref{eq:attractiveness}.
        
        \begin{equation}
        \label{eq:attractiveness}
            A_i^t=\begin{cases}
                1-\frac{\left |N_i^t\bigcap \left(\bigcup_{j\in T-\{t\}}N_i^{j} \right)\right|}{\left | \bigcup_{j\in T} N_i^j \right|}, & \text{when }j\geq0\text{ and }N_i^t\neq\emptyset\\ 
                0, & \text{otherwise}. 
            \end{cases}
        \end{equation}
    \end{itemize}
\end{itemize}

\begin{itemize}[leftmargin=*]
        \item \textit{\textbf{Burst (BB) (set $F_b | F_b\subset F$)}}: bursty behavior is defined as temporal non-homogeneous sequence of events~\cite{Karsai2012} and has been characterized by a fat-tailed inter-event time ($\Delta t$) distribution. In one of the bitcoin blockchain attacks (Allinvain Theft~\cite{bitcointalk}), a malicious account generated a large number of transactions to taint the bitcoin platform. Motivated by  this incident, we define four types of bursts (temporal, degree, balance and gasPrice) that occur in the network under consideration. As an account can either be a sender or a receiver, the following burst types are defined for cases \textit{(a)} when the account acts as a sender, \textit{(b)} when the account acts as a receiver, and \textit{(c)} when the account acts as both a sender as well as a receiver. 
    
        \begin{itemize}[leftmargin=*] 
            \item \textit{\textbf{Temporal Burst}}: 
            for an account $i$, non-homogeneous occurrences of events (in our case transactions) lead to some transactions occurring where $\Delta t$ is less than a threshold, $\theta_t^i$, while for other transactions $\Delta t$ is large. 
            If a transaction happens when $\Delta t < \theta_t^i$, we assume that it is a burst. Some burst can be long lived while some burst can be short lived, meaning, some event can happen continuously for long time intervals before going dormant. As features, we identify number of such temporal bursts (\textit{\textbf{numberOfTemporalBursts}}) and the duration of the longest burst (\textit{\textbf{longestBurstDuration}}) for both in and out transactions separately as well. 
    
            \item \textit{\textbf{Degree Burst}}: it has been proven that the degree (also inDegree and outDegree) distribution of the aggregated transactions in blockchain such as Ethereum follows a power-law (fat tailed) distribution~\cite{chen2018} with $\alpha\in[-2.8,-2.6]$. This suggests that many accounts do not transact often while there are very few accounts that act as hubs (for example, exchanges). Nonetheless, when considering the temporal aspects, we believe such behavior also exists where some accounts have a very high degree for some instant while for other instants they have a low degree. Thus, we define a degree burst when at a given instant of time the degree of an account, $i$, is greater than $\theta_d^i$. Similar to the temporal case, for degree bursts we also identify number of degree bursts (\textit{\textbf{numberOfDegreeBursts}}) that happened for an account over time, number of instances where the degree burst happened (\textit{\textbf{numberOfDegreeBurstInstances}}), and the time at which the largest burst of degree happened (\textit{\textbf{largestBurstAt}}). Note that these features except for numberOfDegreeBurstInstances are defined for both in and out transactions separately as well. 
        
            \item \textit{\textbf{Balance Burst}}: in some cases transactions happen from accounts $i$ to account $j$ where the involved sum of crypto-currency was very large (more than a threshold value $\theta_b^i$). For example, some accounts associated to Silk Road~\cite{Spagnuolo2014} or involved in money laundering sometimes transact large sum for illegal activities. Busty behavior of transaction amount could be helpful in identifying potential malicious activities and accounts. Similar to the above cases, for an account $i$, we identify number of unique instances where balance is more than $\theta_b^i$ (\textit{\textbf{numberOfBalanceBurstyInstances}}), and number transactions more than $\theta_b^i$ (\textit{\textbf{numberOfBalanceBursts}}). Note that, we define these factors for both only in and out case.
            
            \item \textit{\textbf{GasPrice Burst}}: As described before, an attacker can put higher gas price (more than a threshold value $\theta_g^i$) to bribe the miner so that the transaction is included in the block. This activity although abnormal is useful in understanding account's behavior. Towards this, similar to previous cases, we define \textit{\textbf{numberOfGasPriceBurstyInstances}} as number of instances where the gasPrice was set more than $\theta_g^i$. This is only defined for in case as gasPrice is only set by the sender.
        \end{itemize}
        
\end{itemize}

Note that features such as in/outDegree, burst, attractiveness are some graph-based temporal features. Besides these features, other graph-based properties that we use as feature includes \textit{\textbf{clustering-coefficient (CC)}}~\cite{Watts1998}. For an account $i$, let $N_i^{t,in}$ be the neighborhood of account $i$ at time $t$ from which the account has received the crypto-currency, $N_i^{t,out}$ be the neighborhood of account $i$ at time $t$ to which the account has paid the crypto-currency. Thus, the total account degree is  $deg_i^{tot}=|N_i^{t,in}|+|N_i^{t, out}|$. Let $N_i^{t,\leftrightarrow}=N_i^{t,in} \bigcap N_i^{t, out}$ and $a_{ir}=1$ if there is a transaction between $i$ and $r$, otherwise $0$. We similarly define $a_{is}$, $a_{ri}$, $a_{si}$, $a_{rs}$, $a_{sr}$. For a directed graph, CC of account $i$ ($CC_i^t$) at time $t$ is defined as equation~\ref{eq:cc}~\cite{Fagiolo2007}.

\begin{equation}
\label{eq:cc}
       CC_i^t=\frac{\sum_r \sum_s (a_{ir}+a_{ri})(a_{is}+a_{si})(a_{sr}+a_{rs})}{2\left[ deg_i^{tot}(deg_i^{tot}-1) - 2 | N_i^{t,\leftrightarrow}|\right ]}.  
\end{equation}

\section{Results and Evaluation}\label{sec:eval}
We evaluate the effectiveness of our method using Ethereum's external transactions data which is publicly available for download using the \textit{Etherscan APIs}~\cite{etherscanApi}. Note that the APIs do not provide any information about the account (such as the name and the account type). Nonetheless, as the hash of the accounts is available, one can check the associated information using the Ethereum Blockchain Explorer~\cite{etherchain}. We perform all our evaluations using Python.

\begin{figure}[!t]
    \centering
    \subfloat[InDegree Distribution][InDegree Distribution]{
        \includegraphics[width=0.9\textwidth]{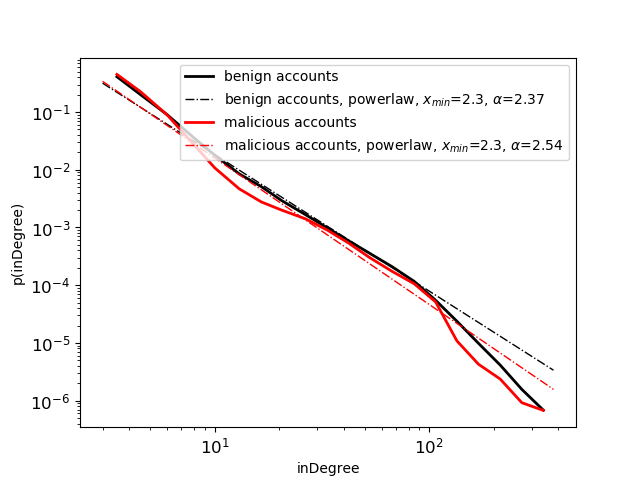}
        \label{fig:indegD}
    }\\
    \subfloat[OutDegree Distribution][OutDegree Distribution]{
        \includegraphics[width=0.9\textwidth]{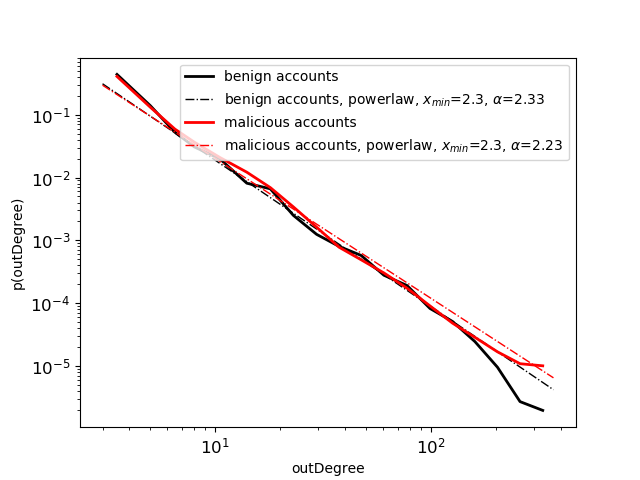}
        \label{fig:outdegD}
    }
    \caption{Degree Distribution of accounts.}
    \label{fig:dd}
\end{figure}

\subsection{Dataset}

Ethereum as on 20th December 2019 had $\approx$79M accounts. Out of these accounts, 3362 accounts were already tagged to be involved in malicious activities. The tags mainly include Phishing (3168 accounts), Gambling (8 accounts), Cryptopia-Hack (6 accounts), Heist (16 accounts), Suspicious (4), Bitpoint Hack (2 accounts), Compromised (21 accounts), Spam (10 accounts), Upbit-Hack (123 accounts), Unsafe (1 account), Scam (1 account) and Bugs (2 accounts). We look for other sources such as Cryptoscam.db~\cite{cryptoscam} as well to know the ground truth about the accounts as some of the accounts might not be tagged as malicious. As a result, we find 329 more malicious accounts with a total of 3691 unique malicious EOAs and SCs. Upon further investigation, we find that out of these 3691 EOAs and SCs, 746 never transacted and were mostly involved in the token trade until 7th December 2019. We thus remove them from our malicious accounts dataset. In these remaining set of malicious accounts, there are 158 SCs and 2 marked compromised exchanges. Note that for these accounts we collect only-but-all external transactions (transactions from EOAs to SCs, and between different EOAs). Also note that at the time of this study Ethereum had removed most of the malicious tags. But recently Ethereum provided new tags and marked more accounts as malicious. As of 27th May 2020, there were 4708 malicious accounts out of which 2019 were newly tagged accounts. Out of these 2019 accounts only 1252 accounts ever transacted. Out of these 1252 accounts 1029 were created before 7th December 2019 in which only 3 are present in our dataset. As the number of malicious accounts is constantly evolving, we take this opportunity to cross validate accounts that our analysis found malicious.

There is a high class imbalance in the dataset as the number of benign accounts is large. Thus, we perform random under-sampling to uniformly sample 697K benign accounts from the 79M Ethereum accounts. In the total $\approx$700K accounts we have, there are 7 exchanges and 23,141 SCs while rest accounts are EOAs.
 
A unique transaction, \textit{\textbf{Tx}}, contains information about blockHash, blockNumber, source, destination, gas, gasPrice, Transaction hash, balance, and timestamp of the block. Note that the Tx data does not include the timestamp of when a transaction was performed by the account. The only time related information, we are able to extract is the information about when a block is mined. However, currently we do not use this information. We assume a time bin of 1 block for our study. We assign respective \textit{blockNumber} as a timestamp to all the transactions\footnote{The block numbers are continuous thus giving a notion of timestamp.}. Based on this notion of timestamp, we also segment the data into several $SD$ of different $T_g$ and study the behavior of the accounts. We describe in the section~\ref{sec:intro} the different $T_g$s we consider. For statistical purposes, we have 1,531 Day $SD$s, 219 Week $SD$s, 52 Month $SD$s, 18 Quarter $SD$s, 9 HalfYearly $SD$s, 5 Year $SD$s, and the entire dataset. A total of 1835 datasets.

For our study we assign: \textit{(i)} $\theta_t$ = 2 so that continuous burst of smallest size are also captured, \textit{(ii)} for an account $i$, $\theta_d^i$ = $0.8\times(max(d))$ where $d$ is the in/outdegree of an account in the considered $SD$, \textit{(iii)} $\theta_b^i= 0.8\times(max(b))$ where $b$ is the transaction balance for either in or out case, \textit{(iv)} $\theta_a^i$ to be equal to the duration of the $SD$ to keep the entire history of neighbors that a particular account transacted in the past in the given that sub-dataset, and \textit{(v)}) $\theta_g^i=0.8\times(max(gasPrice))$ where gasPrice is the the gas price for transactions associated with account $i$. We then analyse different time series based features to identify there characteristics as potential features.

\subsection{Results}
For the entire dataset, we first study inDegree and outDegree distribution for both malicious and benign accounts to validate the fat-tailed behavior of the degree distribution. From fig.~\ref{fig:dd}, we identify that power-law distribution~\cite{Alstott2014} with $x_{min}=2.3$, $\alpha\in[2.37,2.54]$ and $\alpha\in[2.23,2.33]$ fits inDegree and outDegree distribution, respectively, for both malicious and benign accounts. Here $\alpha$ and $x_{min}$ are the powerlaw exponent and minimum  $x$ from where the powerlaw distribution is observed, respectively.

\begin{figure*}[!t]
    \centering
    \subfloat[Temporal inDegree distribution for all accounts][Temporal inDegree distribution for all accounts]{
        \includegraphics[width=0.9\textwidth]{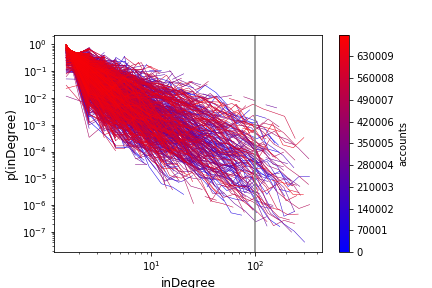}
        \label{fig:indt}
    }\\
    \subfloat[Temporal outDegree distribution for all accounts][Temporal outDegree distribution for all accounts.]{
        \includegraphics[width=0.9\textwidth]{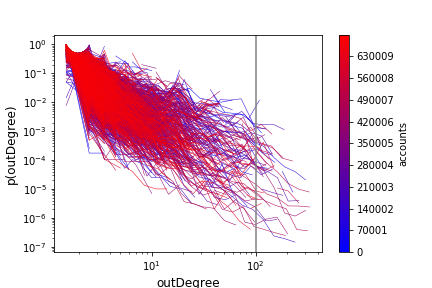}
        \label{fig:outdt}
    }
    \caption{Temporal Degree Distribution of individual accounts.}
    \label{fig:tdd}
\end{figure*}

The fat-tailed nature of degree is evident because some accounts interact with more number of accounts at a certain instant, thereby inducing a bursty behavior. We study the distribution of inDegree for all individual accounts to understand if such behavior is shown by all the accounts. Fig.~\ref{fig:indt} presents distribution of inDegree for different accounts. We identify that the inDegree of very few accounts is high ($>$100) for very few time instances while most of the time it is low suggesting the existence of bursts. We observe a similar behavior for outDegree as well (see fig.~\ref{fig:outdt}). 

\begin{figure*}[!t]
    \centering
    \subfloat[Distribution of $\Delta t$][Distribution of $\Delta t$]{
        \includegraphics[width=0.9\textwidth]{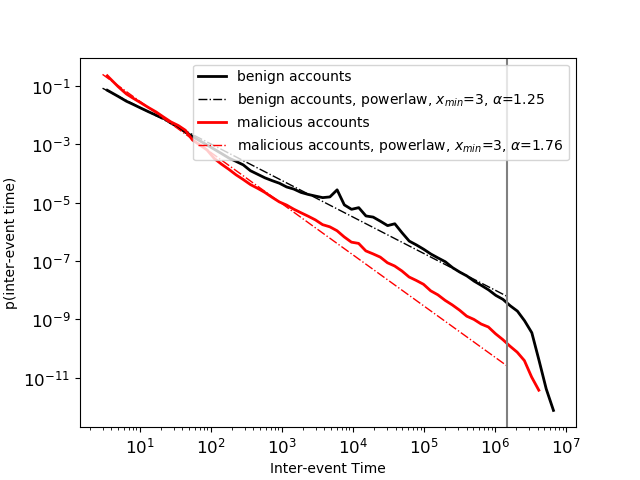}
        \label{fig:ietd}
    }\\
    \subfloat[account-wise distribution of $\Delta t$][account-wise distribution of $\Delta t$]{
        \includegraphics[width=0.9\textwidth]{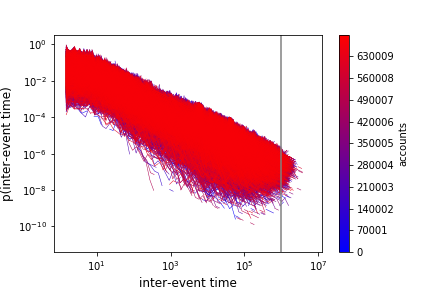}
        \label{fig:aietd}
    }
    \caption{Distribution of $\Delta t$}
    \label{fig:tietd}
\end{figure*}

Next, we validate the existence of temporal bursts. For this we study the distribution of inter-event time ($\Delta t$) for all accounts. We find that it follows power law with $x_{min}=3$ and $\alpha=1.25$ and $\alpha=1.76$ for benign and malicious cases, respectively (see fig.~\ref{fig:ietd}). Nonetheless, we also observe a truncation at $1.5\times10^6$ blocks. The truncation reflects that some accounts are inactive or did not perform any transactions for long period of time. When looking at the individual level, we observe that only few accounts have very large inter-event time ($>1\times10^6$) where the probability of occurrence of such events is very low. Most of the activity happens where the inter-event time is very small (see fig.~\ref{fig:aietd}).

\begin{figure}[!t]
\centering
\includegraphics[width=0.9\textwidth]{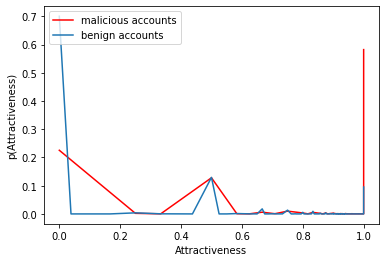}
\caption{Attractiveness}
\label{fig:attractiveness}
\end{figure}

The attractiveness behavior of malicious and benign accounts differ significantly (see fig.~\ref{fig:attractiveness}). Most malicious accounts have high attractiveness value while most of the benign accounts have low attractiveness value. This justifies our assumption that most malicious accounts target those accounts that they have not previously interacted with.Some attacks (Upbit Hack - Fake\_Phishing1431: `0xdf9191889649c442836ef55de5036a7b694115b6') uses multiple accounts to evade detection while transferring money to exchanges. They use multiple accounts as buffer between account and exchange. This is the reason for relatively high probability ($p(A=0)>0.2$) for the low values of attractiveness ($A=0$) for malicious accounts. Similarly for some benign accounts $p(A=1)=0.1$ because such accounts only have 1 incoming transaction in whole lifetime portraying account interacted only with new accounts.

For the entire dataset, after applying tsfresh, for every temporal feature $F_t^j \in F_t$ we get a set of features ($\hat{F}_t^j$) that describes $F_t^j$. From $\hat{F}_t^j$, we choose top three feature. We use \textit{Gini} as the scoring method to identify the top three feature. After this process, we get a total of 59 features. For the entire dataset, using \textit{pearson correlation}, we remove highly correlated features and find 36 important features. We also perform PCA to identify 28 features that cover $>$98.2\% variance to further reduce the feature space in the entire dataset. 

For the analysis purposes, besides performing PCA to identify 28 features and before running the AutoML tool (TPOT) to identify the best supervised learning algorithm, we segment the entire dataset into six dataset configurations. Note that these six dataset configurations are different from the temporal $SD$s. Three out of these six dataset configurations use all types of accounts (EOA and SC) and have 59, 36, and 28 features, respectively. While for the remaining three, we separate EOAs from SCs and use only EOAs. These three configurations again have 59, 36, and 28 features, respectively. We configure TPOT with all the supervised ML algorithms used in the state of the art studies along with other supervised ML algorithms to identify the algorithm that gives best balanced accuracy.

Table~\ref{tab:balAccSup} lists different dataset configurations we have used along with the algorithm that provided the best balanced accuracy along with precision, recall and F1-score for each class. For each dataset configuration and the algorithm that provided the best balanced accuracy, we only provide values to those hyperparameters for which the values are different from the default case. We identify that ExtraTreesClassifier provides overall best balanced accuracy for all the dataset configurations and among them dataset with 59 features and all the account types has best balanced accuracy. The difference in balanced accuracy score between the dataset configurations when 36 and 59 features are used is only 0.5\% for both when we consider only EOAs and all the accounts, respectively. Given such results, we show that correlated features do not provide much gain and can be removed without the loss of accuracy. 

 \begin{figure}[!t]
 \includegraphics[width=\textwidth]{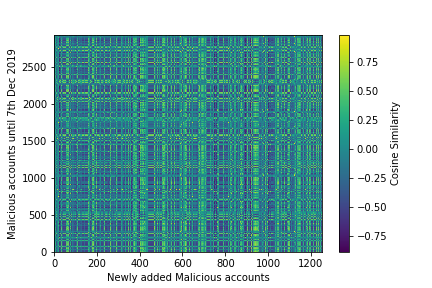}
 \caption{Cosine similarity between newly identified malicious accounts and old malicious accounts.}
 \label{fig:malcs}
 \end{figure}

To validate our results, we test ExtraTreesClassifier with identified hyperparameters on newly identified set of 1252 malicious accounts. The classifier achieves 50\% balanced accuracy. However, when we train the classifier with identified hyperparameters on the total dataset (dataset consisting of previously used 700k accounts and new 1252 accounts), we were able to achieve $\approx92\%$ balanced accuracy. This makes us wonder if the new malicious nodes have different characteristics. We check cosine similarity between the old 2946 malicious accounts and the new 1252 malicious accounts (cf. figure~\ref{fig:malcs}). We find that most of the newly added malicious accounts had low similarity score. Only one new malicious account had similarity score $>0.985$ with only one old malicious account. In many cases the similarity score even reached $<-0.89$ showing that the accounts are not similar and there are some new aspects used by new malicious accounts. Note that to identify cosine similarity we do not use features such as $transactedlast$ and $transactedFirst$ because many of the accounts were created after 7th Dec 2019.

\begin{table}
\centering
\begin{threeparttable}
\begin{tabular}{|c|c|c|c|cc|cc|cc|}
\hline
& & \textbf{TPOT}&  &     &    &     &    &     &  \\
\multirow{2}{*}{\textbf{Features}}& \textbf{Data}& \textbf{identified}& \textbf{Accuracy} 
& \multicolumn{2}{c|}{\textbf{Precision}} & \multicolumn{2}{c|}{\textbf{Recall}} & \multicolumn{2}{c|}{\textbf{F1 score}} \\
\cline{5-10}
    & \textbf{Segment} & \textbf{Classifier}& \multicolumn{1}{c|}{\textbf{balanced}}  & \textbf{Mal}    & \textbf{Ben}   &\textbf{Mal}   & \textbf{Ben}   &  \textbf{Mal}    & \textbf{Ben}  \\
\hline
28 & Only EOA  & ExtraTrees & 0.872 &  0.38 & 1.00  & 0.75 & 0.99 & 0.50 & 1.00 \\
(PCA) & EOA and SC & ExtraTrees  & 0.873 &  0.22  & 1.00  & 0.76  & 0.99  & 0.34  & 0.99\\

\hline
\multirow{2}{*}{36} & Only EOA  & ExtraTrees & 0.876 &  0.11 & 1.00 & 0.78 & 0.97 & 0.19 &0.99 \\
  & EOA and SC & ExtraTrees & 0.882 & 0.24 & 1.00 & 0.78 & 0.99 & 0.37 & 0.99   \\
  
\hline
\multirow{2}{*}{59} 
& Only EOA & ExtraTrees & 0.881 & 0.26 & 1.00 & 0.77 & 0.99 & 0.38 & 0.99\\
& EOA and SC & \textbf{ExtraTrees}  & \textbf{0.887} &  0.29  & 1.00  & 0.78  & 0.99  & 0.42  & 1.00\\
\hline
\end{tabular}
\begin{tablenotes}

\item [28 (PCA) EOA] ExtraTreesClassifier(class\_weight = `balanced', max\_features = 0.4, max\_samples = 0.3, min\_samples\_leaf = 11, min\_samples\_split = 19, n\_estimators = 600)
\item [28 (PCA) EOA and SC] ExtraTreesClassifier(class\_weight = `balanced', criterion = 'entropy', max\_features = 0.25, max\_samples = 0.15, min\_samples\_leaf = 13, min\_samples\_split = 4, n\_estimators = 800, n\_jobs = 20, random\_state = 100)
\item [36 EOA] ExtraTreesClassifier(bootstrap = true, class\_weight = `balanced', max\_features = 0.15, max\_samples = 0.7, min\_samples\_leaf = 8, min\_samples\_split = 18, n\_estimators = 200,  n\_jobs = 10, random\_state = 100)
\item [36 EOA and SC] ExtraTreesClassifier(class\_weight = `balanced', criterion = 'entropy', max\_features = 0.45, max\_samples = 0.75, min\_samples\_leaf = 18, min\_samples\_split = 6, n\_estimators = 200)
\item [59 EOA]  ExtraTreesClassifier(class\_weight = `balanced', max\_features = 0.2, max\_samples = 0.75, min\_samples\_leaf = 13, min\_samples\_split = 19)
\item [59 EOA and SC]  ExtraTreesClassifier(class\_weight = `balanced', criterion = 'entropy', max\_features = 0.3, max\_samples = 0.3, min\_samples\_leaf = 14, min\_samples\_split = 20, n\_estimators = 200)
\end{tablenotes}
\end{threeparttable}
\caption{Balanced accuracy, Precision, Recall and F1 score for both malicious (Mal) and benign (Ben) accounts with best identified ML algorithm for supervised case when using different dataset configurations.}
\label{tab:balAccSup}
\end{table}

\begin{figure}[ht]
\centering
\includegraphics[width=0.9\textwidth]{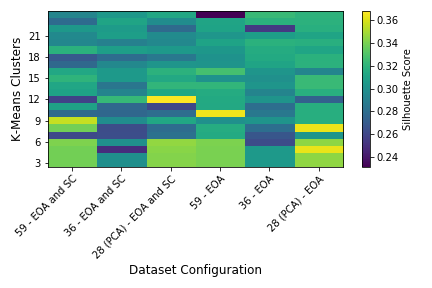}
\caption{Silhouette Scores for clusters identified by K-Means for different dataset configurations and $k\in [3, 24]$.}
\label{fig:silhouette}
\end{figure}

\begin{figure}[!ht]
\centering
\includegraphics[width=0.9\textwidth]{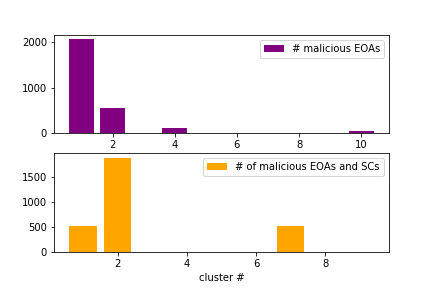}
\caption{Clusters with number of malicious accounts for (a) when only EOAs \\ are considered, (b) when both EOAs and SCs are considered.}
\label{fig:numCluster}
\end{figure}

\begin{figure}[!t]
\centering
\includegraphics[width=0.9\textwidth]{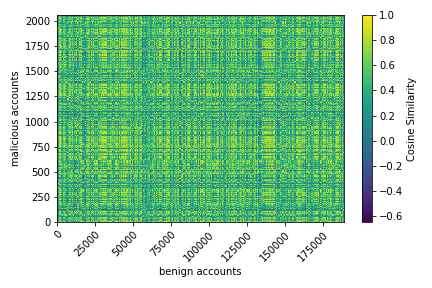}
\caption{Cosine similarity between malicious accounts and benign accounts in \\ the cluster with best Silhouette score.}
\label{fig:cosSim}
\end{figure}

We next test unsupervised learning algorithms such as K-Means, DBSCAN, HDBSCAN, and oneClassSVM to identify suspect accounts in the entire dataset. We find that for the six dataset configurations (mentioned above and not the $SD$s) and different values of $k\in[3,24]$, K-Means provide the best silhouette score (score = 0.365) when $k = 10$ clusters and when we use all the features but only EOAs (`59 - EOA') (see fig.~\ref{fig:silhouette}). Among these 10 clusters, for one initial condition, one cluster had the most number of already known malicious EOAs ($\approx 73.9\%~(2062/2788)$) (see fig.~\ref{fig:numCluster}). We then identify the similarity between all the accounts in the identified cluster. We identify 554 benign accounts whose behavior (cosine similarity) (see fig.~\ref{fig:cosSim}) is within $1-\epsilon$ where $\epsilon\rightarrow 0$ to that of malicious accounts. For our analysis we use $\epsilon=10^{-7}$. We cross validate the transactions performed by these 554 benign accounts and find that \textit{(a)} most of the EOAs have small \textit{transactedLast} value, meaning, those accounts never transacted in recent past (in past 6 months 494 EOAs never interacted), \textit{(b)} atleast 38 EOAs only have incoming transactions and are not exchanges, and \textit{(c)} \textit{totalBalance} $\in [0.0, 150.0]$ Ethers with a median of 0.001 Ethers. 

When considering both EOAs and SCs, we obtain the best silhouette score (score = 0.356) for $k=9$ clusters but for the case when we use all the 59 features (`59 - EOA and SC') (see fig.~\ref{fig:silhouette}). In this case, for one initial condition, there was one cluster with a maximum number of already tagged malicious EOAs ($\approx 64.3\%~(1793/2788)$) and malicious SCs ($\approx 62.6\%~(99/158)$). We identify 293 potential suspects EOAs and no suspect SCs within this cluster using our previous method. Out of these 293 accounts, 160 EOAs were also detected in the set of 554 accounts. We further tested if the accounts we identified as suspects are present in the list of newly tagged malicious accounts. We found that none of the 3 new malicious tagged accounts that transacted during our analysis period were not in our list of suspects. This is possible as the accounts must have changed their behavior and become malicious after our collection period. We do not reveal the account hash for the sake of privacy and not maligning benign accounts in interacting with these either 554 or 293 suspects until they are officially tagged malicious. Other unsupervised ML algorithms did not perform better than K-Means. The range of silhouette scores for HDBSCAN was $\in[-0.06,-0.022]$ while oneClassSVM did not converge. 

To further understand the temporal behavior changes before classifying the accounts as malicious we use temporal sub-datasets ($SD$s) created at different temporal granularities ($T_g$, see section~\ref{sec:intro}). Consider a $T_g\in T_G$ which consists of a several $SD$s. Let this set be set $SD(T_g)$ where $SD(T_g)=\{SD(T_g)_1,~SD(T_g)_2,$ $~\cdots,~SD(T_g)_j,~\cdots,~SD(T_g)_n\}$. Further, consider an account $i$. We first analyse all the time-series based features in each $SD(T_g)_j$ and characterise them. We employ a similar approach as before where we identify $\hat{F}_t^i$ using tsfresh for a $F_t^i\in F$ in a given $SD(T_g)_j$ and use three features in $\hat{F}_t^i$ with highest \textit{gini} score.

\begin{figure}[!t]
\centering
\includegraphics[width=0.9\textwidth]{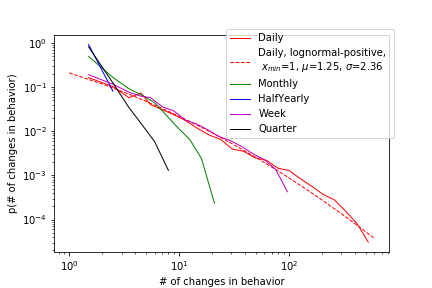}
\caption{Probability distribution of number of changes in behavior of accounts with certain probability for being benign at different $T_g$s.}
\label{fig:changeinBeh}
\end{figure}

We then use K-Means with previously identified hyperparameter ($k=9$) and perform clustering. As before, we tag accounts in each $SD(T_g)_j$ as malicious and benign after identifying cosine similarity. This results in a vector ($M$) for each account of size $n_i$ where each element ($M_{j}$) in $M$ is either 0 or 1 and $n_i$ is the number of $SD$s in a $T_g$ in which the account appears. Here 0 represents not identified as malicious. Let this set of $SD$s be $SD(T_g)=\{SD(T_g)_1^i,~SD(T_g)_2^i,~\cdots,~SD(T_g)_j^i,~\cdots,~SD(T_g)_n^i\}$. $M$ depicts the behavior of an account $i$ where a change in behavior is captured if $M_{j}\neq M_{j+1}$. We note that only one benign account, as per our analysis, has changed its behaviour most number of times (591) in the $T_g=Day$. Figure~\ref{fig:changeinBeh} shows probability distribution of number of changes in behavior performed by accounts. The figure only considers those accounts where the change happened at least once. For the daily case, as the data was significant we identify that lognormal-positive distribution with parameters $x_{min}=1$, $\mu=1.25$, and $\sigma=2.36$ best fits the data. Further, across all $T_g$s there were 9254 unique benign accounts that showed unstable behavior. 

\begin{figure}[!t]
\centering
\includegraphics[width=0.9\textwidth]{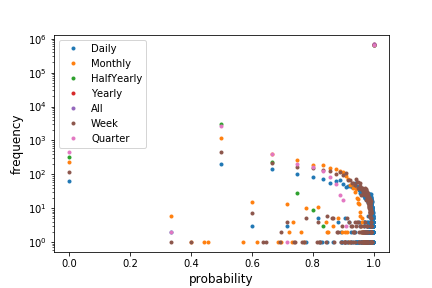}
\caption{Number of accounts with certain probability for being benign at different $T_g$s on a semi-log scale.}
\label{fig:prob}
\end{figure}

From $M$, the probability of a particular account $i$ to be malicious in a given $T_g$ is given by $p_m^i=\frac{\sum_{j\in SD(T_g)^i} M_{j}}{n_i}$. Number of accounts with certain probability for being benign at different $T_g$s is shown in figure~\ref{fig:prob}.  We identify 814 unique accounts across different $T_g$s as suspects that have $p_m^i=0$. Further, as seen from the figure, most of the accounts accounts were identified as benign.

\section{Conclusion}\label{sec:conclusion}

Growth of blockchains technology and concept has found its implementation not only in the financial sector such as crypto-currency market, hedge-fund, and insurance but also in sectors such as governance, education, healthcare, and law enforcement. Although blockchains are privacy-preserving, with an increase in its adoption, security threats are inevitable, more diverse, and deployed using novel techniques. It is essential to have secure transactions. Motivated by the fact that there is limited work in identifying accounts involved in potential malicious activities and those available do not target temporal aspects of blockchains, in this work, we present a way to detect malicious accounts considering the temporal nature of the blockchains.

In this work, we present graph-based temporal features (such as \textit{burst} and \textit{attractiveness}) that are inspired by the existing attacks in the blockchain on top of existing features used to identify malicious accounts. To do so, we first conduct a systematic study of the temporal behavior of the blockchain graph on a collected transaction data in one of the blockchains called Ethereum. Our results show that ExtraTreesClassifier performs best under the supervised setting and achieves balanced accuracy $\in [87.2, 88.7]$ for different dataset configurations. Moreover, under the unsupervised settings, K-Means was able to cluster max 73.9\% known malicious accounts together and identify 554 more suspects that had similar behavior to that of malicious accounts. When considering behavioral changes over time and studying them over different temporal granularities, we are able to detect the probability of an account being malicious at a particular temporal granularity.

Given such results, we expect that benign accounts would be more careful while transacting with suspects and safe-guard themselves from any fraud and security threats. Nonetheless, the current technique is applicable to permissionless blockchain. We would like to investigate the applicability of our method to blockchains where features such as \textit{Transaction Fees} and \textit{Balance} are missing. Despite whether a particular blockchain is permissionless or permissioned, there are many other centrality measures such as closeness, betweenness and page-rank that are applicable in blockchain graph. One another future research direction is to incorporate these measures as features and study the behavior of the accounts before tagging them as malicious or benign. Nonetheless, in this work, we detected suspects using supervised learning and unsupervised learning algorithms. Reinforcement learning is another type of ML that can be applied and studied to detect malicious activity. As our validations failed on the newly tagged malicious accounts one perspective is to study new features and new methods that the new malicious accounts are using and deploying to perform illegal activities.  


\section*{Competing interests}
  The authors declare that they have no competing interests.

\section*{Author's contributions}
    RA, SB and SKS designed the research, RA and SB conducted experiments. All authors read and approved the final manuscript.

\section*{Availability of data and materials}
The list of 2946 malicious accounts used will be made available upon request. Nonetheless, list of all 4708 malicious accounts is publicly available on Etherscan and can be crawled using~\cite{etherscanLC}. The Ethereum transaction Data is also public and can be downloaded using Etherscan APIs~\cite{etherscanApi}.

\section*{Acknowledgements}
  Not applicable.
  
\section*{Funding}
  This work is partially funded by the National Blockchain Project at IIT Kanpur sponsored by the National Cyber Security Coordinator's office of the Government of India and partially by the C3i Center funding from the Science and Engineering Research Board of the Government of India.

\bibliographystyle{bmc-mathphys} 

\bibliography{arxivVersion}

\end{document}